\documentclass{article} 
\usepackage[hidelinks]{hyperref}
\usepackage{url}
\usepackage{graphicx}           
\usepackage{nvmacros}
\usepackage[xindy]{glossaries}
\makeglossaries
\usepackage{nips15submit_e,times}
\nipsfinalcopy
\title{Explaining How a Deep Neural Network Trained with End-to-End
  Learning Steers a Car}
\author{
    Mariusz Bojarski \\
    NVIDIA Corporation \\
    Holmdel, NJ 07733 \\
  \And
    Philip Yeres \\
    NVIDIA Corporation \\
    Holmdel, NJ 07733 \\
  \And
    Anna Choromanaska \\
    New York University \\
    New York, NY 10012 \\
  \And 
    Krzysztof Choromanski \\
    Google Research \\
    New York, NY 10011 \\
  \And
    Bernhard Firner \\
    NVIDIA Corporation \\
    Holmdel, NJ 07733 \\
  \And
    Lawrence Jackel \\
    NVIDIA Corporation \\
    Holmdel, NJ 07733 \\
  \And
    Urs Muller \\
    NVIDIA Corporation \\
    Holmdel, NJ 07733 \\
}
\newacronym{pic}{PIC}{pilot in charge}
\newacronym{ptp}{PTP}{peer-to-peer}
\newacronym{cpu}{GPU}{Graphical Processing Unit}
\newacronym{gtc}{GTC}{\gls{gpu} Technology Conference}
\newacronym{sam}{SAM}{strategic alignment meeting}
\newacronym{pdk}{PDK}{platform development kit}
\newacronym{sdk}{SDK}{software development kit}
\newacronym{por}{POR}{plan of record}
\newacronym{gie}{GIE}{\gls{gpu} Inference Engine}
\newacronym{roi}{ROI}{Region of Interest}
\newacronym{clamp}{CLaMP}{camera lens and mounting parameters}
\newacronym{dave}{DAVE}{DARPA Autonomous Vehicle}
\newacronym{alvinn}{ALVINN}{Autonomous Land Vehicle In a Neural Network}
\newacronym{lrp}{LRP}{layer-wise relevance propagation}
\newacronym{fov}{FOV}{field of view}
\newacronym{dm}{DM}{demosaicing}
\newacronym{nr}{NR}{noise reduction}
\newacronym{adc}{ADC}{analog-to-digital converter}
\newacronym{dlo}{DLO}{digital lateral overflow}
\newacronym{ltm}{LTM}{linear tone mapping}
\newacronym{pwl}{PWL}{piecewise linear}
\newacronym{wdr}{WDR}{wide dynamic range}
\newacronym{bl}{BL}{black-level}
\newacronym{ae}{AE}{auto exposure}
\newacronym{hd}{HD}{high-definition}
\newacronym{fps}{FPS}{frames per second}
\newacronym{cnn}{CNN}{convolutional neural network}
\newacronym{dnn}{DNN}{deep neural network}
\newacronym{nn}{NN}{neural network}
\newacronym{imu}{IMU}{inertial measurement unit}
\newacronym{rl}{RL}{reinforcement learning}
\newacronym{torcs}{TORCS}{the open race car simulator}
\newacronym{gta}{GTA}{Grand Theft Auto}
\newacronym{oem}{OEM}{original equipment manufacturer}
\newacronym{mse}{MSE}{mean squared error}
\newacronym{nyu}{NYU}{New York University}
\newacronym{ue4}{UE4}{Unreal Engine~4}
\newacronym{ai}{AI}{artificial intelligence}
\newacronym{gps}{GPS}{Global Positioning System}
\newacronym{csi}{CSI}{Camera Serial Interface}
\newacronym{usb}{USB}{Universal Serial Bus}
\newacronym{isp}{ISP}{image signal processor}
\newacronym{ros}{ROS}{Robot Operating System}
\newacronym{ilsvrc}{ILSVRC}{Large Scale Visual Recognition Challenge}
\newacronym{can}{CAN}{Controller Area Network}
\newacronym{csv}{CSV}{comma-separated values}
\newacronym{suv}{SUV}{sport utility vehicle}
\newacronym{mph}{MPH}{miles per hour}
\newacronym{darpa}{DARPA}{Defense Advanced Research Projects Agency}
\newacronym{gui}{GUI}{graphical user interface}
\newacronym{convnet}{ConvNet}{convolutional network}
\newacronym{api}{API}{application programming interface}
\newacronym{3d}{3D}{three dimensional}
\newacronym{sota}{SOTA}{state-of-the-art}
\newacronym{dl}{DL}{Deep Learning}
\newacronym{ces}{CES}{Consumer Electronics Show}

\begin{document}
\maketitle
%
%
%
%
%
%
%
%

\begin{abstract}
As part of a complete software stack for autonomous driving, NVIDIA has
created a neural-network-based system, known as {\em PilotNet\/},
which outputs steering angles given images of the road ahead. PilotNet
is trained using road images paired with the steering angles generated
by a human driving a data-collection car. It derives the necessary
domain knowledge by observing human drivers. This eliminates the need
for human engineers to anticipate what is important in an image and
foresee all the necessary rules for safe driving. Road tests
demonstrated that PilotNet can successfully perform lane keeping in a
wide variety of driving conditions, regardless of whether lane
markings are present or not.

The goal of the work described here is to explain what PilotNet learns
and how it makes its decisions. To this end we developed a method for
determining which elements in the road image most influence PilotNet's
steering decision. Results show that PilotNet indeed learns to
recognize relevant objects on the road.

In addition to learning the obvious features such as lane markings,
edges of roads, and other cars, PilotNet learns more subtle features
that would be hard to anticipate and program by engineers, for
example, bushes lining the edge of the road and atypical vehicle
classes.
\end{abstract}

\section{Introduction}
A previous report \cite{dave2-2016} described an end-to-end learning
system for self-driving cars in which a \gls{cnn} \cite{lecun-89e} was
trained to output steering angles given input images of the road
ahead. This system is now called PilotNet. The training data were
images from a front-facing camera in a data collection car coupled
with the time-synchronized steering angle recorded from a human
driver. The motivation for PilotNet was to eliminate the need for
hand-coding rules and instead create a system that learns by
observing. Initial results were encouraging, although major
improvements are required before such a system can drive without the
need for human intervention.  To gain insight into how the learned
system decides what to do, and thus both enable further system
improvements and create trust that the system is paying attention to
the essential cues for safe steering, we developed a simple method for
highlighting those parts of an image that are most salient in
determining steering angles. We call these salient image sections the
{\em salient objects\/}. 
A detailed report describing our saliency detecting method
can be found in \cite{vis-2016}

\begin{figure}[t]
  \hfil
  \includegraphics[scale=0.55]{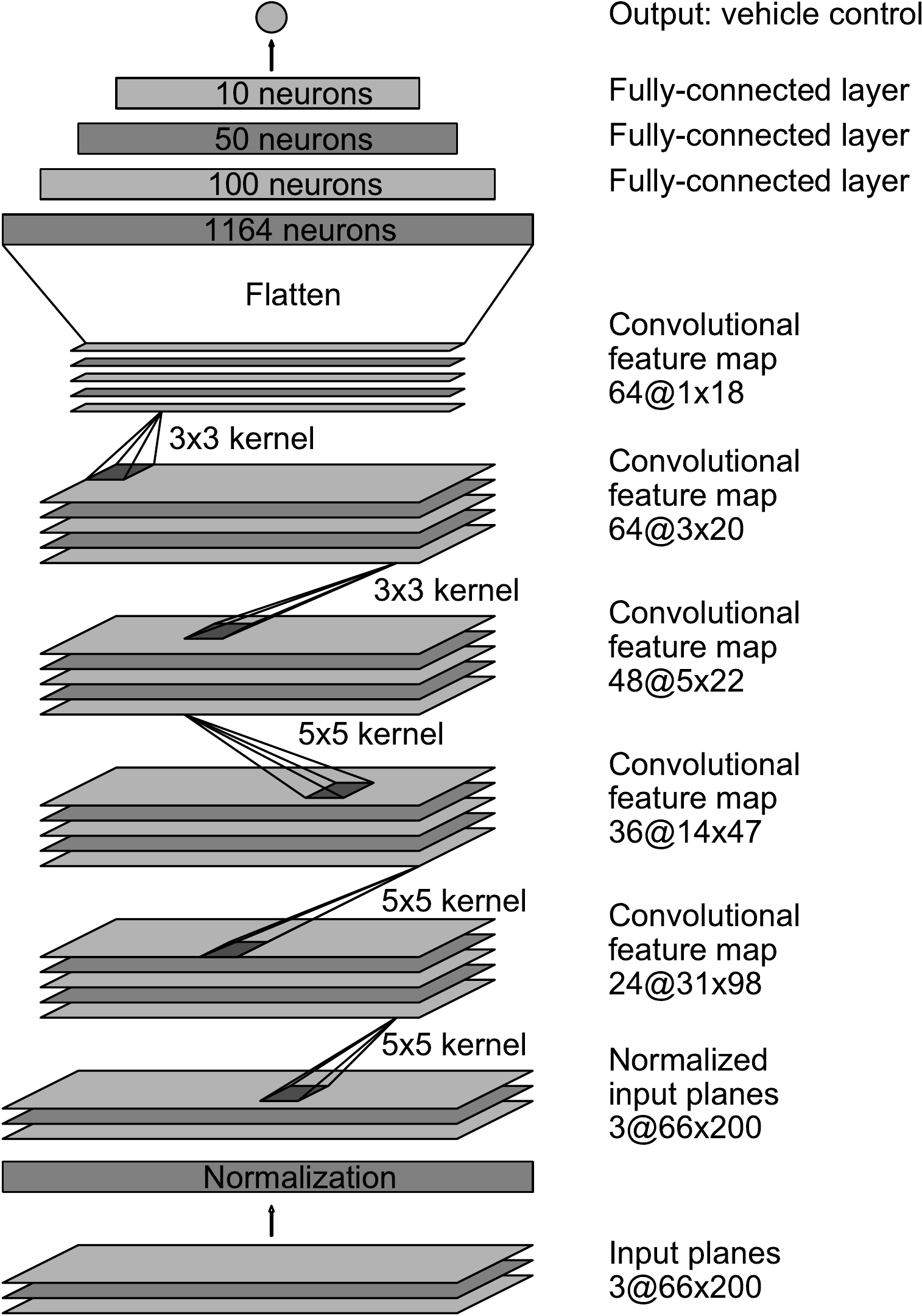}
  \caption{PilotNet architecture.}
  \label{fig-cnn-architecture}
\end{figure}

Several methods for finding saliency have been described by other
authors. Among them are sensitivity based
approaches~\cite{Baehrens:2010:EIC:1756006.1859912,
  DBLP:journals/corr/SimonyanVZ13,Rasmussen_visualizationof},
deconvolution based ones~\cite{MGR11,MR14}, or more complex ones like
\gls{lrp}~\cite{BachPLOS15}. We believe the simplicity of our method,
its fast execution on our test car's NVIDIA
\drivepxtwo\ \glsunset{ai}\gls{ai} car computer, along with its nearly
pixel level resolution, makes it especially advantageous for our task.

\subsection{Training the PilotNet Self-Driving System}
PilotNet training data contains single images sampled from video from
a front-facing camera in the car, paired with the corresponding
steering command ($1/r$), where $r$ is the turning radius of the
vehicle.  The training data is augmented with additional
image/steering-command pairs that simulate the vehicle in different
off-center and off-orientationpoistions. For the augmented images, the
target steering command is appropriately adjusted to one that will
steer the vehicle back to the center of the lane.

Once the network is trained, it can be used to provide the steering
command given a new image.

\section{PilotNet Network Architecture}
The PilotNet architecture is shown in
Figure~\ref{fig-cnn-architecture}.  The network consists of 9 layers,
including a normalization layer, 5 convolutional layers and 3 fully
connected layers. The input image is split into YUV planes and passed
to the network.  The first layer of the network performs image
normalization. The normalizer is hard-coded and is not adjusted in the
learning process.

The convolutional layers were designed to perform feature extraction
and were chosen empirically through a series of experiments that
varied layer configurations. Strided convolutions were used in the
first three convolutional layers with a 2\x2 stride and a 5\x5 kernel
and a non-strided convolution with a 3\x3 kernel size in the last two
convolutional layers.

The five convolutional layers are followed with three fully connected
layers leading to an output control value that is the inverse turning
radius. The fully connected layers are designed to function as a
controller for steering, but note that by training the system
end-to-end, there is no hard boundary between which parts of the
network function primarily as feature extractors and which serve as
the controller.

\begin{figure}[t]
  \hfil
  \includegraphics[width=\textwidth]{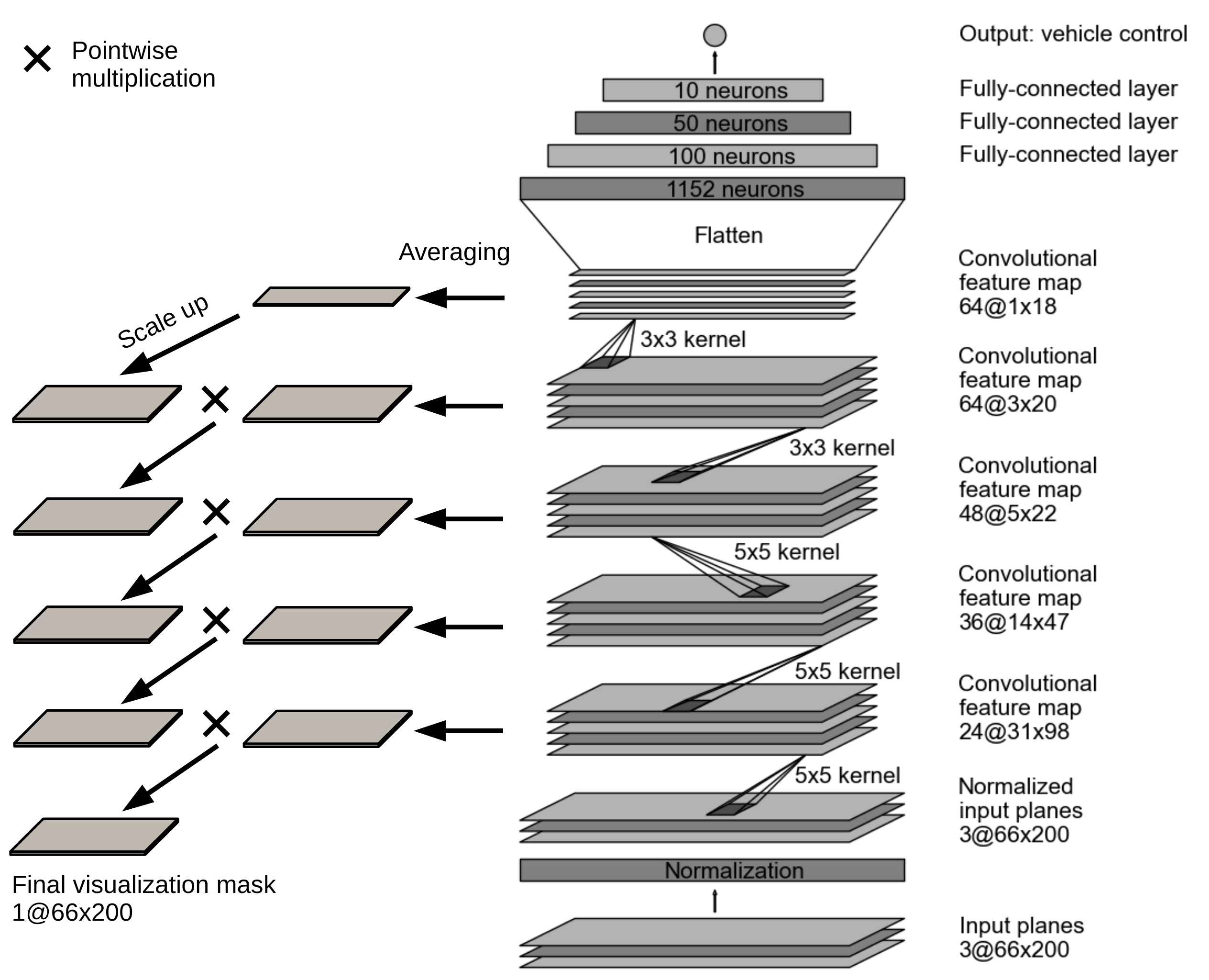}
  \caption{Block diagram of the visualization method that identifies
    the salient objects.}
  \label{fig-block-diagram}
\end{figure}

\section{Finding the Salient Objects}
The central idea in discerning the salient objects is finding parts of
the image that correspond to locations where the feature maps,
described above, have the greatest activations.

The activations of the higher-level maps become masks for the
activations of lower levels using the following algorithm:

\begin{enumerate}
  \item In each layer, the activations of the feature maps are
    averaged.
  \item The top most averaged map is scaled up to the size of the map
    of the layer below. The up-scaling is done using deconvolution.
    The parameters (filter size and stride) used for the deconvolution
    are the same as in the convolutional layer used to generate the
    map. The weights for deconvolution are set to~$1.0$ and biases are set
    to~$0.0$.
  \item The up-scaled averaged map from an upper level is then
    multiplied with the averaged map from the layer below (both are now
    the same size). The result is an intermediate mask.
  \item The intermediate mask is scaled up to the size of the maps of
    layer below in the same way as described Step~2.
  \item The up-scaled intermediate map is again multiplied with the
    averaged map from the layer below (both are now the same size). Thus a new 
    intermediate mask is obtained.
  \item Steps~4 and~5 above are repeated until the input is reached.
    The last mask which is of the size of the input image is
    normalized to the range from $0.0$ to $1.0$ and becomes the final
    visualization mask.
\end{enumerate}

This visualization mask shows which regions of the input image
contribute most to the output of the network.  These regions identify
the salient objects. The algorithm block diagram is shown in
Figure~\ref{fig-block-diagram}.
 
\begin{figure}[p]
  \hfil
  \includegraphics[width=0.4\textwidth]{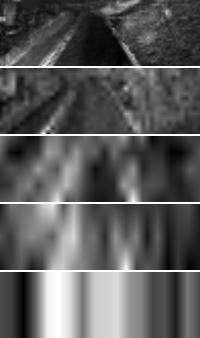}
  \includegraphics[width=0.4\textwidth]{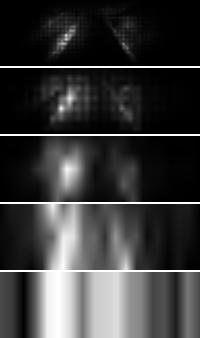}
  \caption{\textbf{Left:} Averaged feature maps for each level of the
    network.  \textbf{Right:} Intermediate visualization mask for each
    level of the network.}
  \label{fig-mask-creation}
\end{figure}

The process of creating the visualization mask is illustrated in
Figure~\ref{fig-mask-creation}. The visualization mask is overlaid on
the input image to highlight the pixels in the original camera image
to illustrate the salient objects.

\begin{figure}[p]
  \hfil\includegraphics[width=0.8\textwidth]{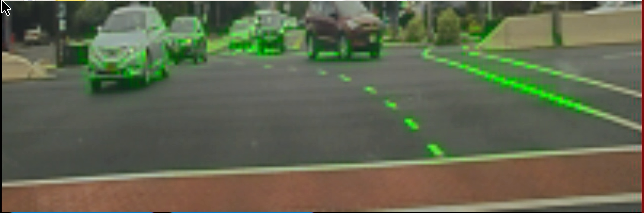}

  \hfil\includegraphics[width=0.8\textwidth]{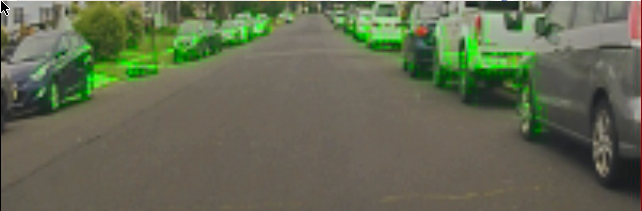}

  \hfil\includegraphics[width=0.8\textwidth]{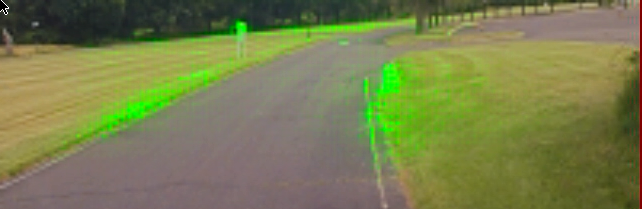}
  \caption{Examples of salient objects for various image inputs.}
  \label{fig-vis}
\end{figure}

Results for various input images are shown in Figure~\ref{fig-vis}.
Notice in the top image the base of cars as well as lines (dashed and
solid) indicating lanes are highlighted, while a nearly horizontal
line from a crosswalk is ignored. In the middle image there are no
lanes painted on the road, but the parked cars, which indicate the
edge of the road, are highlighted.  In the lower image the grass at
the edge of the road is highlighted. Without any coding, these
detections show how PilotNet mirrors the way human drivers would use
these visual cues.

\begin{figure}[htb]
  \hfil\includegraphics[width=0.8\textwidth]{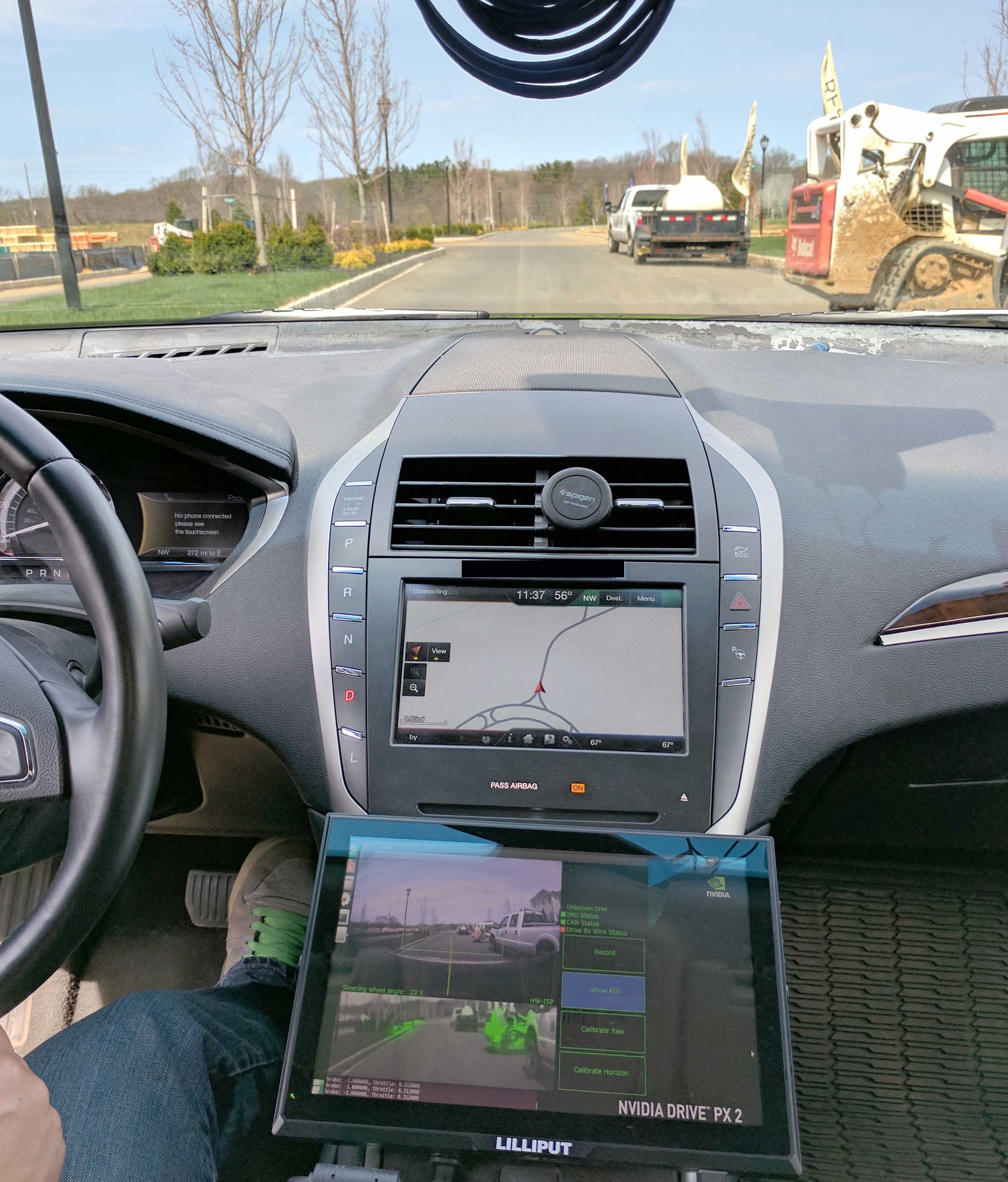}
  \caption{View inside our test car}
  \label{fig-in-car}
\end{figure}

Figure~\ref{fig-in-car} show a view inside our test car. At the top of
the image we see the actual view through the windshield. A PilotNet
monitor is at the bottom center displaying diagnostics.

\begin{figure}[htbp]
  \hfil\includegraphics[width=0.8\textwidth]{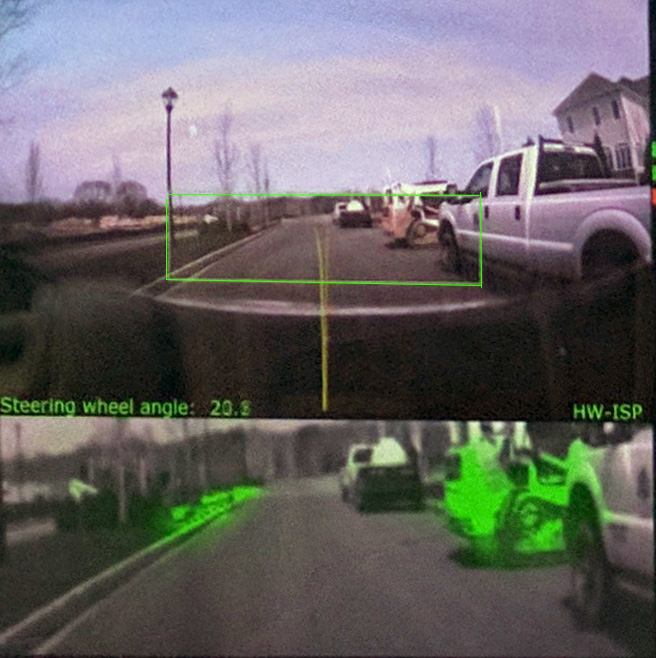}
  \caption{The PilotNet monitor from Figure~\ref{fig-in-car} above}
  \label{blowup}
\end{figure}

Figure~\ref{blowup} is a blowup of the PilotNet monitor.  The top image
is captured by the front-facing camera.  The green rectangle outlines
the section of the camera image that is fed to the neural network.
The bottom image displays the salient regions.  Note that PilotNet
identifies the partially occluded construction vehicle on the right
side of the road as a salient object.  To the best of our knowledge,
such a vehicle, particularly in the pose we see here, was never part
of the PilotNet training data.

\section{Analysis}

While the salient objects found by our method clearly appear to be
ones that should influence steering, we conducted a series of
experiments to validate that these objects actually do control the
steering.  To perform these tests, we segmented the input image that
is presented to PilotNet into two classes.

Class~1 is meant to include all the regions that have a significant
effect on the steering angle output by PilotNet. These regions include
all the pixels that correspond to locations where the visualization
mask is above a threshold.  These regions are then dilated by 30
pixels to counteract the increasing span of the higher-level feature
map layers with respect to the input image. The exact amount of
dilation was determined empirically. The second class includes all
pixels in the original image minus the pixels in Class~1.  If the
objects found by our method indeed dominate control of the output
steering angle, we would expect the following: if we create an image
in which we uniformly translate only the pixels in Class~1 while
maintaining the position of the pixels in Class~2 and use this new
image as input to PilotNet, we would expect a significant change in
the steering angle output.  However, if we instead translate the
pixels in Class~2 while keeping those in Class~1 fixed and feed this
image into PilotNet, then we would expect minimal change in PilotNet's
output.

\begin{figure}[htp]
  \hfil\includegraphics[width=0.9\textwidth]{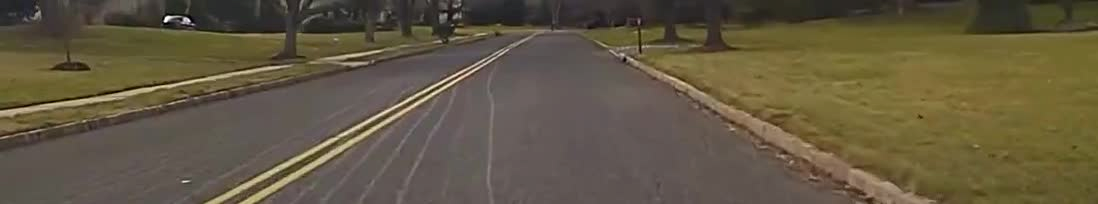}

  \hfil\includegraphics[width=0.9\textwidth]{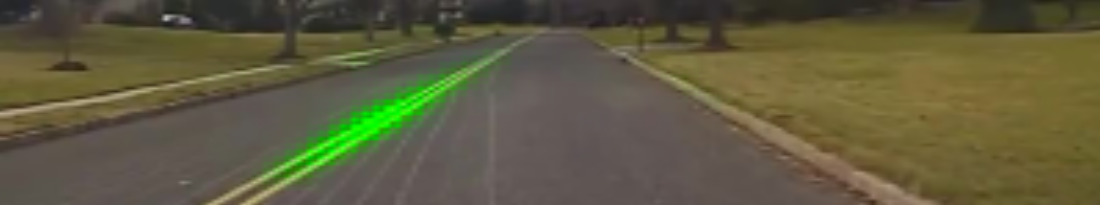}

  \hfil\includegraphics[width=0.9\textwidth]{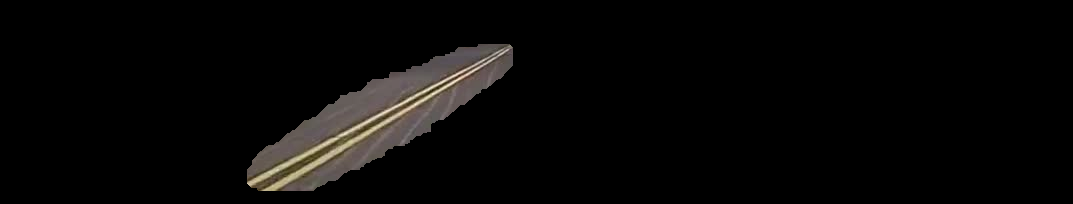}

  \hfil\includegraphics[width=0.9\textwidth]{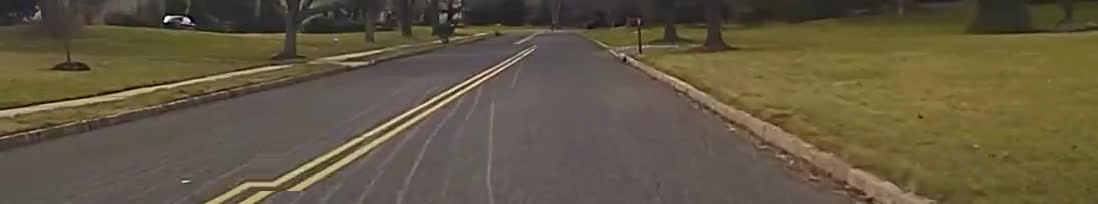}
  \caption{Images used in experiments to show the effect of
    image-shifts on steer angle.}
  \label{fig-roi}
\end{figure}

Figure \ref{fig-roi} illustrates the process described above.  The top
image shows a scene captured by our data collection car.  The next
image shows highlighted salient regions that were identified using the
method of Section~3. The next image shows the salient regions dilated.
The bottom image shows a test image in which the dilated salient
objects are shifted.

\begin{figure}[htp]
  \hfil\includegraphics[width=0.8\textwidth]{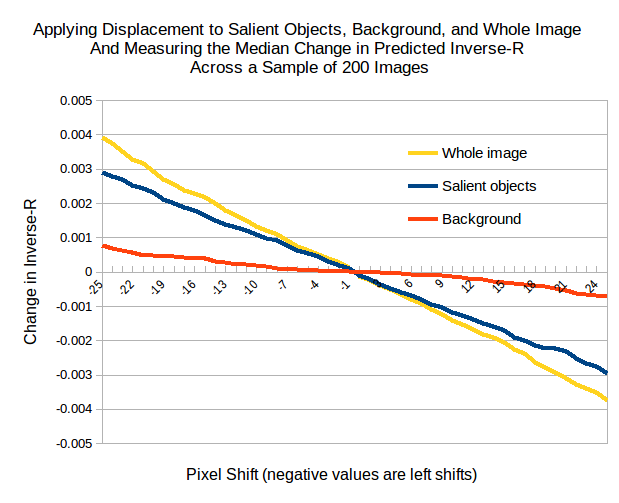}

  \caption{Plots of PilotNet steering output as a function of pixel
    shift in the input image.}
  \label{fig-vis4}
\end{figure}

The above predictions are indeed born out by our experiments.  Figure
\ref{fig-vis4} shows plots of PilotNet steering output as a function
of pixel shift in the input image. The blue line shows the results
when we shift the pixels that include the salient objects (Class~1).
The red line shows the results when we shift the pixels not included
in the salient objects. The yellow line shows the result when we shift
all the pixels in the input image.

Shifting the salient objects results in a linear change in steering
angle that is nearly as large as that which occurs when we shift the
entire image.  Shifting just the background pixels has a much smaller
effect on the steering angle. We are thus confident that our method
does indeed find the most important regions in the image for
determining steering.

\section{Conclusions}
We describe a method for finding the regions in input images by which
PilotNet makes its steering decisions, \ie, the salient objects. We
further provide evidence that the salient objects identified by this
method are correct. The results substantially contribute to our
understanding of what PilotNet learns.

Examination of the salient objects shows that PilotNet learns features
that ``make sense'' to a human, while ignoring structures in the
camera images that are not relevant to driving.  This capability is
derived from data without the need of hand-crafted rules. In fact,
PilotNet learns to recognize subtle features which would be hard to
anticipate and program by human engineers, such as bushes lining the
edge of the road and atypical vehicle classes.

%
\small
\bibliography{neural-nets}

\begin{thebibliography}{1}

\bibitem{dave2-2016}
Mariusz Bojarski, Davide~Del Testa, Daniel Dworakowski, Bernhard Firner, Beat
  Flepp, Prasoon Goyal, Lawrence~D. Jackel, Mathew Monfort, Urs Muller, Jiakai
  Zhang, Xin Zhang, Jake Zhao, and Karol Zieba.
\newblock End to end learning for self-driving cars, April 25 2016.
\newblock URL: \url{http://arxiv.org/abs/1604.07316}, \href
  {http://arxiv.org/abs/arXiv:1604.07316} {\path{arXiv:arXiv:1604.07316}}.

\bibitem{lecun-89e}
Y.~LeCun, B.~Boser, J.~S. Denker, D.~Henderson, R.~E. Howard, W.~Hubbard, and
  L.~D. Jackel.
\newblock Backpropagation applied to handwritten zip code recognition.
\newblock {\em Neural Computation}, 1(4):541--551, Winter 1989.
\newblock URL: \url{http://yann.lecun.org/exdb/publis/pdf/lecun-89e.pdf}.

\bibitem{vis-2016}
Mariusz Bojarski, Anna Choromanska, Krzysztof Choromanski, Bernhard Firner,
  Larry Jackel, Urs Muller, and Karol Zieba.
\newblock {VisualBackProp}: visualizing {CNNs} for autonomous driving, November
  16 2016.
\newblock URL: \url{https://arxiv.org/abs/1611.05418}, \href
  {http://arxiv.org/abs/arXiv:1611.05418} {\path{arXiv:arXiv:1611.05418}}.

\bibitem{Baehrens:2010:EIC:1756006.1859912}
D.~Baehrens, T.~Schroeter, S.~Harmeling, M.i Kawanabe, K.~Hansen, and K.-R.
  M\"{u}ller.
\newblock How to explain individual classification decisions.
\newblock {\em J. Mach. Learn. Res.}, 11:1803--1831, 2010.

\bibitem{DBLP:journals/corr/SimonyanVZ13}
K.~Simonyan, A.~Vedaldi, and A.~Zisserman.
\newblock Deep inside convolutional networks: Visualising image classification
  models and saliency maps.
\newblock In {\em Workshop Proc. ICLR}, 2014.

\bibitem{Rasmussen_visualizationof}
P.~M. Rasmussen, T.~Schmah, K.~H. Madsen, T.~E. Lund, S.~C. Strother, and L.~K.
  Hansen.
\newblock Visualization of nonlinear classification models in neuroimaging -
  signed sensitivity maps.
\newblock {\em BIOSIGNALS}, pages 254--263, 2012.

\bibitem{MGR11}
M.~D. Zeiler, G.~W. Taylor, and R.~Fergus.
\newblock Adaptive deconvolutional networks for mid and high level feature
  learning.
\newblock In {\em ICCV}, 2011.

\bibitem{MR14}
M.~D. Zeiler and R.~Fergus.
\newblock Visualizing and understanding convolutional networks.
\newblock In {\em ECCV}, 2014.

\bibitem{BachPLOS15}
S.~Bach, A.~Binder, G.~Montavon, F.~Klauschen, K.-R. M{\"u}ller, and W~Samek.
\newblock On pixel-wise explanations for non-linear classifier decisions by
  layer-wise relevance propagation.
\newblock {\em PLOS ONE}, 10(7):e0130140, 2015.
\newblock URL: \url{http://dx.doi.org/10.1371/journal.pone.0130140}.

\end{thebibliography}
\bibliographystyle{unsrturl}
%
\label{lastpage}
\end{document}